\pdfoutput=1
\documentclass[11pt]{article}

\usepackage[]{acl}

\usepackage{times}
\usepackage{latexsym}
\usepackage{booktabs}

\usepackage[T1]{fontenc}
\usepackage[utf8]{inputenc}

\usepackage{microtype}

\usepackage{graphicx}
\usepackage{comment}
\usepackage{bm}
\usepackage{amsmath}
\usepackage{footmisc}
\usepackage{amssymb}

\title{Simple and Effective Knowledge-Driven
Query Expansion\\ for QA-Based Product Attribute Extraction}

\author{Keiji Shinzato\\
Rakuten Institute of Technology,\\
Rakuten Group Inc.\\
\texttt{keiji.shinzato@rakuten.com}
\And Naoki Yoshinaga\\
Institute of Industrial Science,\\
the University of Tokyo\\
\texttt{ynaga@iis.u-tokyo.ac.jp}
\AND Yandi Xia\\
Rakuten Institute of Technology,\\
Rakuten Group Inc.\\
\texttt{yandi.xia@rakuten.com}
\And Wei-Te Chen\\
Rakuten Institute of Technology,\\
Rakuten Group Inc.\\
\texttt{weite.chen@rakuten.com}}
\usepackage{multirow}
\newcommand{\yn}[1]{\textcolor{blue}{#1}}
\newcommand{\kj}[1]{#1}

\newcommand{\pos}[1]{\textcolor[rgb]{0,0,0.85}{#1}}
\newcommand{\nega}[1]{\textcolor[rgb]{0.75,0,0}{#1}}

\makeatletter
\newcommand\footnoteref[1]{\protected@xdef\@thefnmark{\ref{#1}}\@footnotemark}
\makeatother

\begin{document}
\maketitle
\begin{abstract}
% \yn{To equip e-commerce sites with faceted search functionality and better product recommendations},
% \yn{consumers of e-commerce sites with a better search experience}, 
% ynaga ここだけ受動態なのがきもい; product / values に例を入れても良い
% For e-commerce sites, attribute extraction from product titles and descriptions is essential technique to associate products with their attribute values. 
A key challenge in attribute value extraction (\textsc{ave}) from e-commerce sites
is how to handle 
a large number of attributes for diverse products.
Although this challenge is partially addressed by a question answering (\textsc{qa}) approach
which finds a value in product data for a given query (attribute),
% where attributes are queries
% ; for a given query (attribute) and context (product text), \kj{models} extract an answer (value of the attribute) from the context. 
% ynaga ここ、伝わらないと思うけどどうすればいいか
% 本当は、表現学習により、クエリとなる属性間の類似性を考慮することで、属性を跨いで学習データを共用できるということまで言えると良いけど、スペース的に入らない
% Although this approach can handle various attributes,
% by learning their representations and capturing their similarities
it does not work effectively for rare and ambiguous queries.
% , or inappropriate
% attributes.
% since their names are affected by 
% named according to 
% various business factors.
% ynaga
% In real-world applications, some attributes are not optimal expressions due to the fact that they are named according to business factors.
%
We thus propose simple knowledge-driven query expansion based on possible answers (values) of a query (attribute) for \textsc{qa}-based % product 
\textsc{ave}. We retrieve values of a query (attribute) from the training data to expand the query.
% to augment the query. 
We train a model with two tricks, knowledge dropout and knowledge token mixing, which mimic the imperfection of the value knowledge in testing. Experimental results on our cleaned version of AliExpress dataset show that our method improves the performance of \textsc{ave} (+6.08 macro F$_1$),
% the query with the values shows better performance 
% than one without the values, 
especially for rare and ambiguous attributes  
% inappropriate 
(+7.82 and +6.86 macro F$_1$, respectively).
% when the attributes with improper notation.

\end{abstract}

\section{Introduction}

One of the most challenging problems 
in attribute value extraction (\textsc{ave}) from e-commerce sites is a data sparseness problem caused by the diversity of attributes.\footnote{AliExpress.com classifies products
in the Sports \& Entertainment category using 77,699 attributes~\cite{xu_2019}.}
%
% Moreover, models based on \textsc{ner} have to introduce one set of chunk tags (e.g., BIO) for each class, and this makes the number of labels that models are required to predict a few times larger than the number of classes.
% In real-world applications,
% It is hard to solve the \textsc{ave} task with tens of thousands of labels using \textsc{ner}-style sequence labeling.
% \yn{The traditional approach such as named entity recognition works \kj{pooly} for \textsc{ave} from e-commerce data \kj{because} it is difficult to collect massive labeled data for 
% minor and 
% rare attributes (\S~\ref{sec:rels}).
To alleviate the data sparseness problem, recent researches~\cite{xu_2019,wang_2020} formalize the task as question answering (\textsc{qa})
%, or more concretely machine reading comprehension, 
to exploit the similarity of attributes via representation learning.
% Over the past two years, approaches that formalize the \textsc{ave} task as Machine Reading Comprehension (\textsc{mrc}) are proposed to overcome the large attribute size problem~\cite{xu_2019,wang_2020}.
% We refer it Attribute Comprehension (AC) in this paper.
% In those approaches, attributes such as brands and colors are regarded as a {\it query} in \textsc{mrc} while product texts such as titles and descriptions are regarded as {\it contexts}.
Specifically, the \textsc{qa}-based \textsc{ave} takes an attribute name as \textit{query} and product data as \textit{context}, and attempts to extract the value from the context. Although this 
% \textsc{qa}-based 
approach mitigates the data sparseness problem,
% in information extraction, 
performance depends on the quality of query representations~\cite{li_2020}. Because attribute names are short and ambiguous
% (\textit{e.g.}, ``function''), 
%  and ``feature''
% or inappropriate 
as queries, % since their names are affected by
% chosen according to
%various 
%business factors, 
the extraction performance drops significantly for rare attributes with
% inappropriate
ambiguous names (\textit{e.g.}, \textit{sort}) which do not represent their values well.
% ynaga; 以下はイントロのストーリーの理解には不要
% The input to the models is a pair of attribute name and product texts, and the output is beginning and ending positions of a span in the product texts of an attribute value corresponding to the given attribute. In the \textsc{mrc} framework, the number of labels that the models need to predict is always fixed in two because attributes for which the models extract values are given as inputs.
% Therefore, the number of labels that the models predict is at most a few. For instance, in case of the BIO tags, it is always three labels no matter how much the number of attributes increases.

\begin{figure}[t]
    \centering
    \includegraphics[width=\columnwidth]{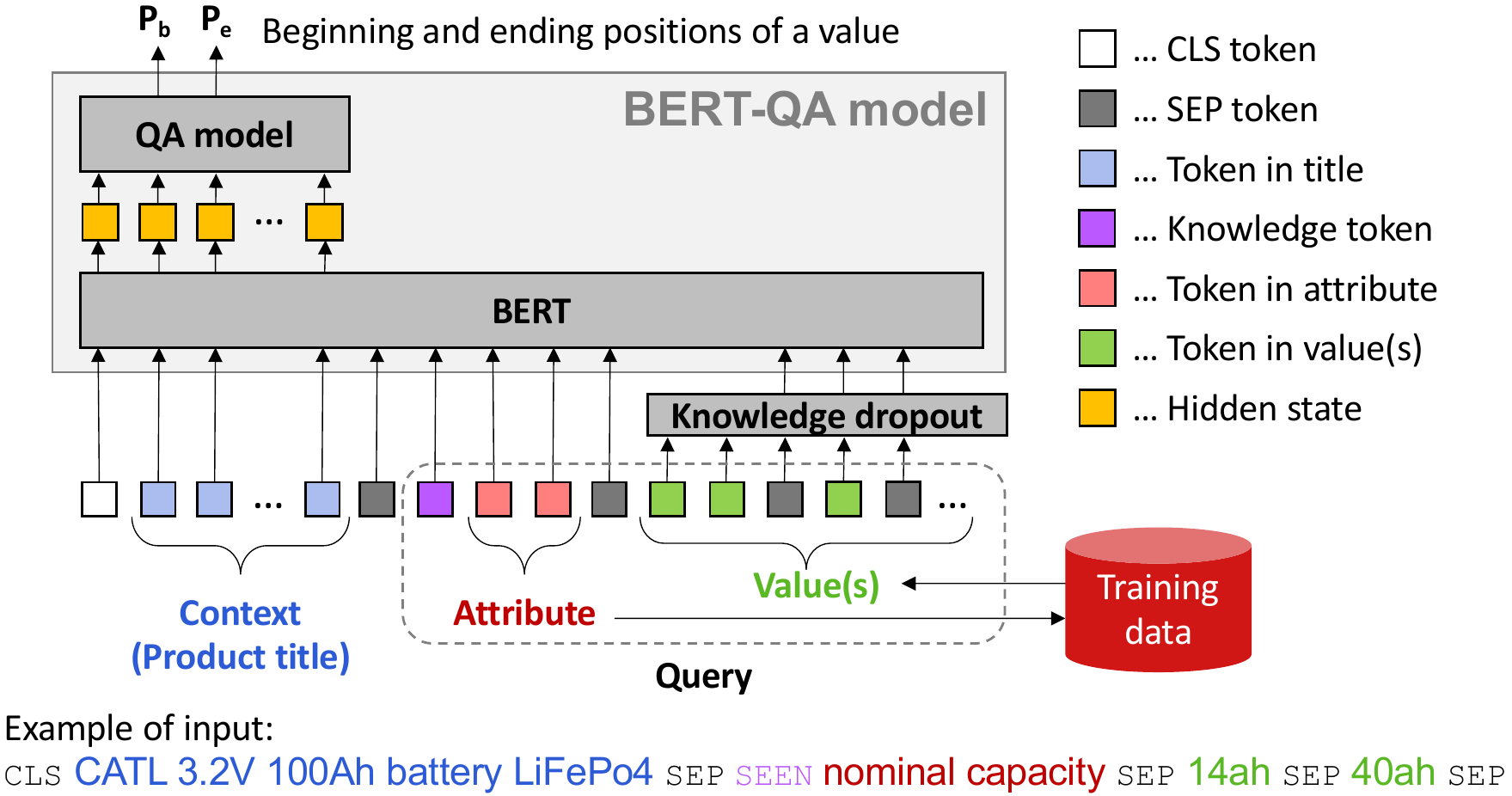}
    \caption{Our knowledge-based \textsc{bert-qa} model for attribute value extraction.}
    \label{fig:model}
\end{figure}
    % \textsc{qa} (\textsc{aveqa}) model~\cite{wang_2020}. 
    % with our query expansion. 
    % The model 
    % It retrieves values for a given attribute from the training data to expand the query.}
% ynaga
\begin{comment}
When we apply \textsc{mrc}-based approaches in real-world applications, attributes are picked up among ones that are being used in actual services. Those attributes, however, are not optimal expressions sometimes due to the fact that they are named according to business factors, or are supposed to be displayed with other information in product pages.
For example, there are cases where an attribute name is an expression such as ``function'' and ``features'' whose specific value is difficult to recall just by looking at the notation.
Li~et~al.~\cite{li_2020} reported that the performance of \textsc{ner} systems based on \textsc{mrc} depends on a query given to the systems. Therefore, we can expect to improve the performance of the models if we can reformulate those attributes to ones that properly express values.
\end{comment}

% In this paper, we propose an \textsc{mrc}-based model for the \textsc{ave} task.
Aiming to perform more accurate
% informative queries for 
\textsc{qa}-based \textsc{ave} for rare and ambiguous 
attributes, % (\S~\ref{ssec:qaave}),
we propose simple query expansion that exploits values for the attribute as knowledge to learn better query representations  (Figure~\ref{fig:model}, \S~\ref{sec:proposed}). We first retrieve possible values of each attribute from the training data, and then use 
% all
the obtained values to augment the query (attribute). 
% To effectively utilize diverse values for the input attribute, 
Since unseen values and attributes will appear in evaluation, we apply dropout to the seen values to mimic the incompleteness of the knowledge (\S~\ref{ssec:vdrop}), and perform multi-domain learning to capture the absence of the knowledge (\S~\ref{ssec:ktm}).
% that can prevent the model from excessively relying on the observed values.}
% explore various strategies to summarize the values for the query.
% Although existing \textsc{mrc}-based models use only attributes as a query, our model exploits attribute values in addition to the attributes in order to enhance the query representation.
%
% This is based on the assumption that by appending an attribute value to a query, we can more precisely and detailedly let the model know what query is than we give the attribute alone because the value is an instance of the attribute.
% In a neural network-based model, the conversion of attribute expression can be regarded as the conversion of embedding representation for attributes.
% Our model employs \textsc{bert}~\cite{devlin_2019} to encode attributes and values to embedding representation, and converts the representation for the attributes using ones for values.

We demonstrate the effectiveness of the query expansion for \textsc{bert}-based \textsc{ave} model~\cite{wang_2020} using the AliExpress dataset\footnote{\label{note1}\url{https://github.com/lanmanok/ACL19_Scaling_Up_Open_Tagging}} released by~\citet{xu_2019}
% https://raw.githubusercontent.com/lanmanok/ACL19_Scaling_Up_Open_Tagging/master/publish\_data.txt
(\S~\ref{sec:experiments}).
In the \kj{evaluation process}, we found near-duplicated data in this dataset.
% that we can easily detect by a pre-processing such as removing extra-whitespaces and lowercasing. We 
We thus % cleaned up the original dataset to 
construct, from this dataset, a more reliable dataset called cleaned \textsc{ae}-pub  to evaluate our method.
% , and then evaluate our method on this cleaned dataset.
%
% The experimental results show that models with the query based on both attribute and value outperform a model without the attribute value, and that the query with the value is more effective for attributes that do not have a proper notation.
%
\begin{comment}
\begin{table}[t]
    \centering
    \caption{Comparison between sequence labeling methods based on \textsc{mrc}. $\textsc{bert}_{\mathit{emb}}$ denotes the embedding layer in \textsc{bert}, and $f$ denotes the function returning an annotation guideline note for the class $c$.}
    \begin{tabular}{p{0.195\columnwidth}|p{0.17\columnwidth}|p{0.23\columnwidth}|p{0.26\columnwidth}}\toprule
    Domain & Model & Query & How to get query representation\\\midrule
    E-commerce & SU OpenTag~\cite{xu_2019} & Attribute $a$& BiLSTM(\textsc{bert}($a$)) \\\cmidrule{2-4}
     & \textsc{aveqa}~\cite{wang_2020} & Attribute $a$& $\textsc{bert}_{\mathit{emb}}(a)$ \\\cmidrule{2-4}
     & Ours & Attribute $a$ and it's value $v$ & $\textsc{bert}_{\mathit{emb}}(a, v)$ \\\midrule
     %$\mathit{Average}(\mathit{Attention}(E_{\mathit{\textsc{bert}}}(a \oplus v_i)))$\\\hline
%$\displaystyle
% \frac{1}{|V_a|}{\sum_{i=1}^{|V_a|}}e_{a,v_i}^{1} \oplus \cdots \oplus \frac{1}{|V_a|}{\sum_{i=1}^{|V_a|}}e_{a,v_i}^{L}$\\\hline
    General and biomedical & \textsc{bert-mrc}~\cite{li_2020} & Annotation guideline note of class $c$ & $\textsc{bert}_{\mathit{emb}}(f(c))$\\\bottomrule
    \end{tabular}
    \label{tab:comparison_with_existing_work}
\end{table}
\end{comment}

Our contribution is threefold: \begin{itemize}
    \item We proposed knowledge-driven query expansion for \textsc{qa}-based \textsc{ave} (\S~\ref{sec:proposed});
    % which advances the % performance of the state-of-the-art
    % ~\cite{wang_2020} (\S~\ref{comparison}).
    the knowledge taken from the training data is valuable (\S~\ref{comparison}).
\item
We revealed that
% on the cleaned \textsc{ae}-pub dataset that 
 % attributes with 
% inappropriate 
rare, ambiguous attributes deteriorate the performance of \textsc{qa}-based \textsc{ave} in the e-commerce domain (\S~\ref{comparison}).
\item
% We resolved the quality issues of the 
% public
% AliExpress dataset (\S~\ref{ssec:dataset}), % To facilitate more reliable evaluation, 
% and will release our cleaned dataset. 
%  with the training, dev, and test split.
We will release our cleaned version of AliExpress dataset for research purposes.
\end{itemize}

\begin{comment}
The contributions of our work are as follows:
\begin{itemize}
    \item We propose a query reformulation for \textsc{qa}-based \textsc{ave} (\S~\ref{sec:proposed}), which \yn{advances the state-of-the-art \textsc{ave} performance of the \textsc{qa}-based \textsc{ave} model}~\cite{wang_2020}
    % the \textsc{ave} performance for rare attributes with inappropriate names
    (\S~\ref{comparison}). 
    \item \yn{We confirm on the cleaned \textsc{ae}-pub dataset that rare attributes with ambiguous
    % inappropriate 
    names deteriorate the performance of \textsc{qa}-based \textsc{ave} in the e-commerce domain (\S~\ref{comparison}).}
    % As far as we know, this is a first work to leverage attribute values as a query in \textsc{qa}-based \textsc{ave} models.
    % ynaga
    \item We point out the quality issues of the public AliExpress dataset (\S~\ref{ssec:dataset}). To facilitate more reliable evaluation, we will release our cleaned dataset with the training/dev/test split.
    % \item We show that a public dataset for the \textsc{ave} task contains near-duplicated data that we can easily detect by a pre-processing. In addition, the dataset was not pre-partitioned for training, development, and test data, and previous works using this dataset~\cite{xu_2019,wang_2020} did not specifically mention which data were for training, development and test data. To facilitate more reliable evaluation and comparison on the \textsc{ave} task, we will release our cleaned dataset after partitioning it to training, development, and test sets.
\end{itemize}
\end{comment}

\section{Related Work}
\label{sec:rels}

\begin{comment}

Early attempts to attribute value extraction used unsupervised or minimally-supervised methods based on 
% lexico-syntactic 
patterns or layouts on the Web such as tables~\cite{chen_2000,yoshida_2003,yoshinaga_2006,dalvi_2009,gulhane_2010,ferrez_2012}.
%
To achieve higher precision and recall, researchers then migrated to supervised methods based on named entity recognition (\textsc{ner}) to extract attribute values from text
% product data (product titles and descriptions) 
and disambiguate their types (attributes)~\cite{putthividhya_2011,shinzato_2013,more_2016,zheng_2018,rezk_2019,karamanolakis_2020,dong_2020,zhu_2020,mehta_2021,jain_2021,yan_2021}.
\end{comment}

Attribute value extraction has been modeled as a sequence labeling problem~\cite{putthividhya_2011,shinzato_2013,more_2016,zheng_2018,rezk_2019,karamanolakis_2020,dong_2020,zhu_2020,mehta_2021,jain_2021,yan_2021}.
%traditional approach to extract attribute values from product text use named entity recognition (\textsc{ner})
% methods since it can be regarded as \textsc{ner} on e-commerce domain
% product data (product titles and descriptions) 
% and disambiguate their types (attributes)
%There are many attempts based on \textsc{ner} techniques on the \textsc{ave} task~
% These methods need to introduce a set of chunk tags for each attribute.
%
However, since the number of attributes can \kj{exceed ten thousand} in e-commerce sites,  
%the number of labels to be predicted also gets very large. Thus, 
% we cannot expect that 
% \textsc{ner}
% -based methods 
the models perform poorly for % a wide variety of
the majority of attributes that rarely appear in the labeled data~\cite{xu_2019}. % in real-world applications

% \yn{To alleviate the data sparseness problem, researchers have started to utilize models for question answering (\textsc{qa})\kj{,} or more specifically machine reading comprehension\kj{,} to turn the task to find spans of values for input attribute from the text~\cite{xu_2019,wang_2020}.}
% ynaga 一旦コメントアウト
%Table~\ref{tab:comparison_with_existing_work} shows the summary of \textsc{mrc}-based approaches that tackle the \textsc{ave} and \textsc{ner} tasks.
%
To alleviate the data sparseness problem, \citet{xu_2019} 
% first 
introduced a \textsc{qa}-based approach for the \textsc{ave} task. It separately encodes product titles and attributes using \textsc{bert}~\cite{devlin_2019} and \kj{bi-}directional long-short term memory~\cite{sepp_1997}, and then combines the resulting vectors via an attention layer 
to learn spans of values
%that the model needs to extract
for the attributes from the titles.
\citet{wang_2020} proposed a purely \textsc{bert}-based model\kj{, which} feeds a string concatenating the given title and attribute to \textsc{bert}\@.
% The association between the attribute and spans (values) to be extracted in the context is learned through the Transformer~\cite{vaswani_2017} layers.
%
% In addition to the \textsc{ave} task, the model is trained on the no-answer classification task, and the distilled masked language model task by multitask learning so that the model generalization ability is enhanced.
%They reported that their \textsc{qa}-model with the multitask learning showed slightly lower overall performance than the vanilla \textsc{qa}-model while the multitask learning was effective to extract values for attributes not in the training data (zero-shot extraction).
% They demonstrated that their method outperformed one proposed by Xu~et~al. on a publicly available data set.
%
% Although they have also performed multitask learning to address \kj{unseen attributes} in the training data, \kj{the overall performance} became slightly worse.
% did not improve
% \citet{yan_2021} builds in-house datasets on a few attributes to explore
These \textsc{qa}-based \textsc{ave} models, however, do not fully enjoy the advantage of the \textsc{qa} model, since attribute queries are much shorter than sentential questions in the original \textsc{qa} task.
% These \textsc{qa}-based \textsc{ave} models, however, still suffer from the data sparseness problem in learning query representations. % \kj{This is because they only use} attributes as queries, 
% This is critical since attributes (queries) are much shorter than question sentences in the original \textsc{qa} task.
%In these previous works, an attribute is only fed to the models as a query. On the other hand, our method feeds an attribute value to the model in addition to an attribute. By using the attribute value, we expect that we can more precisely convey what kinds of values needs to be extracted to the model.

To build better queries in solving named entity recognition via \textsc{qa}, \citet{li_2020} exploited 
% discussed the importance of designing informative queries in \textsc{qa}-based \textsc{ner}, and 
annotation guideline notes for 
% individual 
named entity classes as queries.
%(\textit{e.g.}, ``\textit{Find organizations including companies, agencies and institutions.}'' for \textit{organization} class) 
% such as \textit{organization} and \textit{person}, 
% and use them as queries instead of the class labels.
% is converted to a corresponding annotation guideline note.
%, which is the guideline provided to annotators of the data set by the data set builder. 
%For example, in case of the CoNLL2003 dataset~\cite{tjong_2003} where four classes of entities are annotated, the class \textit{organization} is converted to the note ``Find organizations including companies, agencies and institutions.'' 
% Through experiments on several \textsc{ner} data sets, 
% They reported that 
% queries based on
% converting the class names to 
% the annotation guideline notes 
% \yn{the desriptive queries} yield better \textsc{ner} performance.
% is more effective than simply feeding the class names to the \textsc{mrc} model.
%
% The idea of reformulating class names is the same with ours. In real-world applications, however, it is very costly to prepare such detailed guideline note to more than ten thousands of attributes.
Although this approach will be also effective for \textsc{qa}-based \textsc{ave}, it requires substantial labors to prepare manual annotations for more than \kj{ten thousand attributes in e-commerce site.}
% We thus need a lightweight query reformulation for \textsc{qa}-based \textsc{ave}.

\section{Proposed Method}
\label{sec:proposed}

% In this study, we propose 
This section proposes
a simple but effective query expansion method for \textsc{qa}-based \textsc{ave}~\cite{wang_2020} % ,li_2020
% in the training data of \textsc{qa}-based \textsc{ave}\@.
by utilizing attribute values.
Given a product data (title) $\bm{x} = \{x_1,...,x_n\}$ and an attribute $\bm{a} = \{a_{1},...,a_{m}\}$,
%and an attribute value $\bm{v} = \{v_{1},v_{2},...,v_{l}\}$,
where $n$ and $m$ denote the number of tokens, % in the title and attribute, respectively, 
the model 
% is required to 
returns \kj{the} beginning position, $P_b$, and ending position, $P_e$, of a value. % corresponding to attribute $\bm{a}$ in the title $\bm{x}$, where $P_b \leq P_e$.

Figure~\ref{fig:model} depicts the model architecture with our approach.
Although our query expansion is essentially 
applicable to any \textsc{qa}-based \textsc{ave} models, we here employ the state-of-the-art
% \textsc{qa}-based \textsc{ave} 
model \kj{using} \textsc{bert} proposed by~\citet{wang_2020}. % We follow Wang~et~al.~\cite{wang_2020} and Li~et~al.~\cite{li_2020}, we treat the \textsc{ave} task as a \textsc{qa} task.
In addition to the \textsc{qa} component for \textsc{ave}, their model has other two components; the no-answer classifier and the distilled masked language model. % The former is for explicitly handling the missing value cases and the latter is for improving the model generalization ability. They perform the multitask learning to train the unified model. 
Since those components
% are designed to properly extract values for unseen attributes, they 
slightly decrease the overall micro F$_1$,
% for seen attributes. Thus, 
we employ the \textsc{qa} component from their model (hearafter, referred to as \textsc{bert-qa}).

% \yn{Figure~\ref{fig:model} depicts the model architecture with our query reformulation method.} % Although the core architecture is the same as a \textsc{bert}-based \textsc{qa} model that achieved the best overall F$_1$ score~\cite{wang_2020}, 
% Our query reformulation method \kj{utilizes} values observed in the training data in addition to the input attribute to better describe the query.}
% Since various values will be allowed for the input attributes, we explore various ways to encode the observed values for the input attributes.
% We explore the query reformulation based on the attribute value. 
%In what follows, we first explain the \textsc{bert}-based \textsc{qa} model for the \textsc{ave} task (\S~\ref{ssec:qaave}), and then present our value-driven query reformulation methods for \textsc{qa}-based \textsc{ave} models (\S~\ref{sec:proposed}).
%

\subsection{Knowledge-Driven Query Expansion for \textsc{qa}-Based \textsc{ave}}
\label{ssec:idea}
% As mentioned, 
It is inherently difficult for \textsc{qa}-based \textsc{ave} models to induce effective query representations for rare attributes with 
ambiguous
% inappropriate 
names. It is also hard to develop expensive resources such as annotation guideline notes~\cite{li_2020}
for more than ten thousand of attributes in e-commerce domain.

Then, is there any low-cost resource (knowledge) we can leverage to understand attributes? Our answer to this question is values (answers) for the attributes; we can guess what attributes means from their values. %  (instances)
In this study, we exploit attribute values retrieved from the training data\footnote{We can 
% of course
% exploit values obtained from any resources for the query reformulation. For example,
utilize, if any, external resources for our method. For example, e-commerce sites may develop attribute-value databases to organize products in the marketplace.} of the target \textsc{ave} model
% which are promising resources.} 
as \textit{run-time knowledge} to induce better query representations.

% An issue to be addressed in utilizing attribute values for query expansion is how to effectively exploit multiple values for the attribute.
\begin{comment}
% values because the attributes have multiple values in general.
Thus, we explore the following 
% value-driven 
query expansion 
strategy for \textsc{qa}-based \textsc{ave}
to argument the query representation, 
and empirically compare their performance in experiments (\S~\ref{sec:experiments}). 
\end{comment}
%
% Once a value $\bm{v}$ is obtained, we create the concatenated string $[\textsc{cls};\bm{x};\textsc{sep};\bm{a};\textsc{sep};\bm{v}]$, and then feed the string to the \textsc{qa}-based \textsc{ave} model to predict the beginning and ending positions of a value in the title for the given attribute.
%
% The most naive approach % to exploit the value for the query reformulation 
% is to
% For this issue, 
Our query expansion allows the \textsc{qa}-based \textsc{ave} model, $M_{\textsc{qa}}$, to utilize the seen values for attribute $\bm{a}$ in the whole training data to find beginning and ending positions of a value, $\langle P_b, P_e\rangle$ in title $\bm{x}$:
\begin{align}
\langle P_b, P_e\rangle & = M_{\textsc{qa}} ([\textsc{cls};\bm{x};\textsc{sep};\bm{a};\textsc{sep};\bm{v_a}])
\end{align}
Here, % $M_{\textsc{qa}}$ is a \textsc{qa}-based \textsc{ave} model,
% $[\bm{s};\bm{t}]$ represents a concatenated string of $\bm{s}$ and $\bm{t}$,
$\textsc{cls}$ and $\textsc{sep}$ are special tokens to represent a classifier token and a separator, respectively,
and $\bm{v_a}$ is a string concatenating the seen values of the attribute $\bm{a}$ with \textsc{sep}
% This forms the string $[\bm{v_1};\textsc{sep};...;\textsc{sep};\bm{v_k}]$.
%
% We create the string $[\textsc{cls};\bm{x};\textsc{sep};\bm{a};\textsc{sep};\bm{v}]$ by concatenating the title $\bm{x}$, attribute $\bm{a}$ and values $\bm{v}$, and then feed the string to the \textsc{qa}-based \textsc{ave} model.
%
% We then use the concatenated values for the query reformulation in the same way of treating a single value.
% Here,
% We 
% sort the values 
% assume that the values are 
% concatenated in the order of their occurrence 
% sorted 
in descending order of frequency in the training data. %  prior to the concatenation.}
%
% If the input token sequence is longer than the max token length of \textsc{bert}, we truncate the string $\bm{v_a}$.

\subsection{Knowledge Dropout}
\label{ssec:vdrop}
By taking all the seen values in the training data to augment input queries, 
% there is possibility of overfitting such that
the model may
% based on the approach
just learn to match the seen values with one in the given title.
To avoid this, inspired from word dropout employed in language modeling~\cite{gal_2016},
% inspired by~\citet{gal_2016}, 
we perform \textit{knowledge dropout} over $\bm{v_a}$ in training before concatenating it with title $\bm{x}$ and attribute $\bm{a}$.
\begin{align}
\bm{v_a} & = [\textrm{drop}(\bm{v_{a,1}});\textsc{sep};\textrm{drop}(\bm{v_{a,2}});\textsc{sep};\ldots]
\end{align}
Here, drop is a function that replaces a value $\bm{v_{a,i}}$ in $\bm{v_a}$ with padding tokens according to a dropout rate;
% If the value consists of multiple tokens, 
we replace each token in $\bm{v_{a,i}}$ with \textsc{pad}.

To decide if the dropout applies to a value, we take account of the number of examples labeled with the value.
% where the value is labeled in the training data. 
Given the dropout rate $r$ and the number of training examples $n_v$, the dropout performs over the value $\bm{v}$ according to the probability of $r^{n_v}$.
This implementation
% mimics the incompleteness of knowledge more naturally;
captures the fact that infrequent values are more likely to be unseen.

\subsection{Knowledge Token Mixing}
\label{ssec:ktm}
Since values are literally valuable to interpret attributes, the \textsc{qa}-based \textsc{ave} model may rely more on values than an attribute name. This will hurt the performance on unseen attributes whose values are not available. To avoid this, 
% inspired from  multi-domain learning, 
% Since we collect attribute values for the query expansion from the whole training data, the model can refer values for all attributes during training. In the testing phase, however, the model requires to extract values for \textit{unseen} attributes as well as seen attributes, and the model cannot refer any values in case of the unseen attributes. This is different from what the model observed during training, and thus it will hurt the performance. It is necessary to deal with seen and unseen attributes separately.
%
% Moreover, even within seen attributes, there are two types of attributes; attributes whose value are in context and those whose value are not. The reason why we meet the latter is that there is a limit to the number of tokens we can feed to \textsc{bert}. As mentioned, we truncate an input token sequence if it exceeds the max token length of \textsc{bert}. Because of this, the query expansion cannot make use of all the values for attributes that have a large number of values such as brands. It would be better to separately deal with such attributes.
% To distinguish seen and unseen attributes, we employ \textit{domain token mixing} proposed in neural machine translation~\cite{britz_2017}. 
we assume the availability of value knowledge to be \textit{domain}, and perform multi-domain learning for \textsc{qa}-based model with and without our value-based query expansion. This will allow the model to handle not only seen attributes but also unseen attributes. % \kjj{We expect that this can control the degree of dependency on values.}

Inspired from domain token mixing~\cite{britz_2017}, we introduce two special domain tokens (\textit{knowledge tokens}), and prepend either of the tokens to the attribute to express the knowledge status: \textsc{seen} and \textsc{unseen} (with and without values).\footnote{The original domain token mixing learns to induce domain tokens prior to generating outputs, whereas we prepend domain tokens to inputs since the knowledge status is known.} 
%
% Due to no unseen attributes when training the model, 
% We artificially generate unseen attributes from the seen attributes by not expanding the query with the attribute values. 
In training, from an example with title $\bm{x}$ and attribute $\bm{a}$, we build [\textsc{cls};$\bm{x}$;\textsc{sep};\textsc{seen};$\bm{a}$;\textsc{sep};$\bm{v_a}$] and [\textsc{cls};$\bm{x}$;\textsc{sep};\textsc{unseen};$\bm{a}$;\textsc{sep}], and then put these examples to the same mini-batch.
% feed each to the model as the query.
%
In testing, we use \textsc{seen} and \textsc{unseen} tokens for seen attributes (with values) and unseen attributes, respectively.

\section{Experiments}
\label{sec:experiments}
We evaluate our query expansion method for \textsc{qa}-based \textsc{ave} on a public dataset,\footnoteref{note1} which is 
% selected and 
built from product data under the Sports \& Entertainment category in AliExpress, following~\cite{wang_2020}.

\subsection{Settings}
\paragraph{Dataset}
\label{ssec:dataset}
The public AliExpress dataset consists of 110,484 tuples of $\langle$product title, attribute, value$\rangle$. When a value of the attribute is % not presented in 
absent from the title, the value in the tuple is set as ``\textsc{null}.''
% We refer this dataset as \textit{original dataset}.
We manually inspected the tuples in the dataset, and found quality issues;
% that prevent a reliable evaluation;
some tuples contained \textsc{html} entities, and extra white spaces in titles, attributes, and values, and the same attributes sometimes have different letter cases.
% had differences in case despite being the same attribute.
% ynaga tone down
% Although the \textsc{bert} tokenizer of the current \textsc{qa}-based \textsc{ave} model will not be affected by these fluctuations, it will introduce a reproducibility issue. 
% Furthermore, when normalizing white spaces and letter cases, we found \yn{736} duplicated tuples.
% which prevent reliable evaluation.
% ynaga
% ちょっと短く言いたい
%We therefore normalize white spaces and letter case; we 
We thus decoded \textsc{html} entities, converted trailing spaces into a single space, and removed white spaces at the beginning and ending. We also normalized the attributes by putting a space between alphabets and numbers and by removing `:' at the endings (from `feature1:' to `feature 1').
% We therefore removed the extra white spaces\footnote{More precisely, we remove white spaces from beginning and endings, and convert a sequence of white spaces into a single white space.}, and lowercased the titles, attributes, and values. Furthermore, for attributes, we removed `:' if it appears at the end of the attribute name to cancel difference such as ``feature:'' and ``feature'', and insert a white space before digits to fill the gap such as ``feature1'' and ``feature 1.'' We think that these pre-processings do not affect the meaning of titles, attributes, and values. After that, we deduplicated tuples to evaluate models more reliably. 
As \kj{a result}, we found 736 duplicated tuples. By removing these duplicated tuples, 
we finally obtained the \textit{cleaned \textsc{ae}-pub} dataset of 109,748 tuples with 2,162 unique attributes and 11,955 unique values.
% The number of tuples with the value ``\textsc{null}'' is 21,557.
% We refer to this dataset as \textit{cleaned \textsc{ae}-pub dataset} in the experiments.
We split this dataset into training, development, and test sets with the ratio of 7:1:2  (Table~\ref{tab:data_stats}). 
% The statistics of each set are listed in Table~\ref{tab:data_stats}.
% The amount of training and test sets used by Wang et al.~\cite{wang_2020} is also shown for reference. The training and test sets in the cleaned \textsc{ae}-pub dataset are smaller than the datasets used by Wang et al. because of deduplicating the tuples and creating the development set.

\begin{table}[t]
\small
    \centering
    \begin{tabular}{l@{\quad}r@{\;\;}r@{\;\;}r@{\;\;}r}\toprule
     & \textrm{Train} & \textrm{Dev.} & \textrm{Test} \\\midrule
    \# of tuples  & 76,823 & 10,975 & 21,950\\
    \# of tuples with ``\textsc{\textsc{null}}''  & 15,097 & 2,201 & 4,259\\
    \# of unique attribute-value pairs & 11,819 & 2,680 & 4,431\\
    \# of unique attributes & 1,801 & 635 & 872\\
    \# of unique values & 9,317 & 2,258 & 3,671\\\midrule
    \# of tuples \cite{wang_2020} & 88,479 & N/A & 22,005\\\bottomrule
    \end{tabular}
    \caption{Statistics of the cleaned \textsc{ae}-pub dataset.}
    \label{tab:data_stats}
\end{table}

\paragraph{Evaluation Metrics}
We use precision (P), recall (R) and F$_1$ score as 
% evaluation
metrics.
% Following previous studies~\cite{xu_2019,wang_2020}, we
We adopt exact match criteria~\cite{xu_2019} in which the full sequence of extracted value needs to be correct.
%
% We train the models five times with varying random seeds, and average the results.
% All results are reported as the average scores of four trials.
% To remove the influence of random seeds, we train models four times with different seeds, and then report the average of each metric.
% For tuples with ``\textsc{null}'' value, we judge as correct if the model does not extract any values.
% There is a possibility that the model extracts attribute values redundantly if the same expression appears repeatedly in a title. Hence, we deduplicate extraction results before computing the evaluation metrics.

\subsection{Models}

We apply our knowledge-driven query expansion method (\S~\ref{sec:proposed}) to \textsc{bert-qa}~\cite{wang_2020}, a \textsc{qa}-based \textsc{ave} model on \textsc{bert}.
% ~\cite{devlin_2019}
% to evaluate the impact on the \textsc{ave} task.
%
To perform the query expansion, we simply collect values other than ``\textsc{null}'' from tuples in the training data for each attribute (Table~\ref{tab:data_stats}).
% When answering a query (attribute) whose values are not seen in the training data (400 test examples for 260 attributes), the \textsc{bert-qa} is used instead.

%
For comparison, we use
SUOpenTag~\cite{xu_2019},
% We reimplemented both of them because their codes to train models were not  publicly available.
\textsc{aveqa}
%\footnote{The loss  of the distilled masked language model got NaN if we followed the algorithm in the paper.
% to obtain word distributions.
%We instead used BERTMLMHead class implemented in Transformers.\footnoteref{note2}}
%~\cite{wolf_2020}, 
 and vanilla \textsc{bert-qa}~\cite{wang_2020}, which
%
% \citet{wang_2020} reported that \textsc{aveqa} 
achieved the state-of-the-art micro F$_1$ score on the AliExpress dataset.
% \textsc{aveqa} is one of the strongest baselines at the moment.
%
%
% The dictionary-based approach is the simplest idea that exploits values in the training data.
% In addition, we also use a dictionary-based approach that refers attribute-value pairs retrieved for the query expansion. When multiple values match a title, the approach selects the most frequent value in the training data. Similar with our methods, we used the vanilla \textsc{aveqa} for attributes out of the training data.
We also perform a simple dictionary matching; % using seen values for a given attribute. 
it returns the most frequent seen value for a given attribute among those included in the given title.

\begin{table*}[t]
\small
    \centering
    \begin{tabular}{@{\,\,}lc@{\quad}c@{\quad}cc@{\quad}c@{\quad}c@{\,\,}}\toprule
\multirow{2}{*}{Models} & \multicolumn{3}{c}{Macro} &  \multicolumn{3}{c}{Micro}\\
%\cmidrule(lr){2-4}
%\cmidrule(l){5-7}
& P (\%) & R (\%) & F$_1$ & P (\%) & R (\%) & F$_1$\\\midrule
Dictionary &
% 72.96 & 74.07 & 73.51 & 39.15 & 36.34 & 37.69\\
% 39.86 \scriptsize{($\pm$0.62)} & 37.10 \scriptsize{($\pm$0.62)} & 38.43 \scriptsize{($\pm$0.62)} &
% 72.96 \scriptsize{($\pm$0.04)} & 74.10 \scriptsize{($\pm$0.03)} & 73.53 \scriptsize{($\pm$0.04)} \\ 

33.20 \scriptsize{($\pm$0.00)} & 30.37 \scriptsize{($\pm$0.00)} & 31.72 \scriptsize{($\pm$0.00)} &
73.39 \scriptsize{($\pm$0.00)} & 73.77 \scriptsize{($\pm$0.00)} & 73.58 \scriptsize{($\pm$0.00)} \\ 

SUOpenTag~\cite{xu_2019} &
% 86.70 & 79.04 & 82.69 & 30.76 & 27.86 & 29.24\\
30.92 \scriptsize{($\pm$1.44)} & 28.04 \scriptsize{($\pm$1.48)} & 29.41 \scriptsize{($\pm$1.44)} &
86.53 \scriptsize{($\pm$0.78)} & 79.11 \scriptsize{($\pm$0.35)} & 82.65 \scriptsize{($\pm$0.20)} \\

\textsc{aveqa}~\cite{wang_2020} &
41.93 \scriptsize{($\pm$1.05)} & 39.65 \scriptsize{($\pm$0.96)} & 40.76 \scriptsize{($\pm$0.98)} &
86.95 \scriptsize{($\pm$0.27)} & 81.99 \scriptsize{($\pm$0.13)} & 84.40 \scriptsize{($\pm$0.09)} \\ 

\textsc{bert-qa}~\cite{wang_2020} &
42.77 \scriptsize{($\pm$0.36)} & 40.85 \scriptsize{($\pm$0.22)} & 41.79 \scriptsize{($\pm$0.28)} & 
87.14 \scriptsize{($\pm$0.54)} & 82.16 \scriptsize{($\pm$0.21)} & 84.58 \scriptsize{($\pm$0.24)} \\ \midrule

\textsc{bert-qa}  +vals &
39.48 \scriptsize{($\pm$0.37)} & 35.60 \scriptsize{($\pm$0.44)} & 37.44 \scriptsize{($\pm$0.38)} & 
\textbf{88.82} \scriptsize{($\pm$0.22)} & 81.77 \scriptsize{($\pm$0.14)} & 85.15 \scriptsize{($\pm$0.14)} \\

\textsc{bert-qa}  +vals +drop &
41.61 \scriptsize{($\pm$0.83)} & 38.22 \scriptsize{($\pm$0.80)} & 39.84 \scriptsize{($\pm$0.81)} &
88.46 \scriptsize{($\pm$0.26)} & 82.02 \scriptsize{($\pm$0.37)} & 85.12 \scriptsize{($\pm$0.14)} \\

\textsc{bert-qa} +vals +mixing & 46.67 \scriptsize{($\pm$0.33)} & 43.32 \scriptsize{($\pm$0.50)} & 44.93 \scriptsize{($\pm$0.39)} & 88.30 \scriptsize{($\pm$0.69)} & 82.46 \scriptsize{($\pm$0.30)} & \textbf{85.28} \scriptsize{($\pm$0.26)} \\

\textsc{bert-qa} +vals +drop +mixing & \textbf{47.74} \scriptsize{($\pm$0.54)} & \textbf{44.82} \scriptsize{($\pm$0.75)} & \textbf{46.23} \scriptsize{($\pm$0.64)} & 87.84 \scriptsize{($\pm$0.39)} & \textbf{82.61} \scriptsize{($\pm$0.07)} & 85.14 \scriptsize{($\pm$0.19)} \\

\bottomrule
% \textsc{ner}-No-value
% & 87.61 & 83.94 & 85.73
% & 50.82 & 48.28 & 49.51\\
% \textsc{ner}-Single$_{\mathit{freq}}$
% & 87.63 & 84.33 & 85.95
% & 58.68 & 55.08 & 56.82\\
% \textsc{ner}-Single$_{\mathit{rand}}$
% & 87.44 & 84.28 & 85.83
% & 57.24 & 53.92 & 55.53\\
% \textsc{ner}-Multi
% & 88.36 & 83.85 & 86.05
% & 56.37 & 51.22 & 53.67\\\hline
% Ours
% & 87.91 & 84.30 & 86.07
% & 58.85 & 55.32 & 57.02\\\hline
    \end{tabular}
    \caption{Performance on the cleaned \textsc{ae}-pub dataset in Table~\ref{tab:data_stats}; reported numbers are mean (std.\ dev.) of five trials.}
    % The second best scores are underlined.
    % We ignored 70 attributes that have only \textsc{null} values in the test data (178 examples) in computing macro metrics, since we cannot compute recall (and F$_1$) for them.
% To see the effectiveness of the query reformulation, we calculate the macro performance based on attributes whose values are observed in the training data and which takes the query reformulation. As a result, 21,550 examples for \yn{612} attributes remain for the evaluation.}
    \label{tab:performance_on_limited_attrs}
\end{table*}
\begin{table*}[t]
%\begin{minipage}[t]{\columnwidth}

\small
    \centering
    \begin{tabular}{l@{\;\;\;}c@{\;\;\;}r@{\,\,}r@{\quad}r@{\,\,}r@{\quad}r@{\,\,}rr@{\,\,}r@{\quad}r@{\,\,}r@{\quad}r@{\,\,}r}\toprule
    & & \multicolumn{6}{c}{Macro} & \multicolumn{6}{c}{Micro}\\
    \multirow{1}{*}{\textrm{Models}} & \multirow{1}{*}{\textrm{cos}} & \multicolumn{6}{c}{\textrm{Number of training examples (median: 8)}} & \multicolumn{6}{c}{\textrm{Number of training examples (median: 8)}}\\
    & & \multicolumn{2}{c}{$[1, 8)$}
    & \multicolumn{2}{c}{$[8, \infty)$}     
    & \multicolumn{2}{c}{all}
    & \multicolumn{2}{c}{$[1, 8)$}
    & \multicolumn{2}{c}{$[8, \infty)$}     
    & \multicolumn{2}{c}{all} \\
    \midrule
    
    \multirow{2}{*}{\textsc{bert-qa}}&
    lo & 42.41 & \pos{({\scriptsize $+$}0.80)} & 57.58 & \pos{({\scriptsize $+$}5.84)} &  49.96 & \pos{({\scriptsize $+$}3.31)} & 55.65 & \pos{({\scriptsize $+$}8.34)} & 78.51 & \pos{({\scriptsize $+$}1.33)} &  77.59 & \pos{({\scriptsize $+$}1.77)}\\
    \multirow{2}{*}{+vals} & hi & 41.05 & \nega{({\scriptsize $-$}1.75)} & 69.02 & \pos{({\scriptsize $+$}1.00)} & 56.58  & \nega{({\scriptsize $-$}0.23)}  & 57.72 & \pos{({\scriptsize $+$}6.04)} & 88.54 & \nega{({\scriptsize $-$}0.06)} & 88.21  & \pos{({\scriptsize $+$}0.05)} \\
    & all & 41.77 & \nega{({\scriptsize $-$}0.40)} & 63.63 & \pos{({\scriptsize $+$}3.29)} & 53.28 & \pos{({\scriptsize $+$}1.55)}  & 56.65 & \pos{({\scriptsize $+$}7.25)} & 86.39 & \pos{({\scriptsize $+$}0.25)} & 85.89 & \pos{({\scriptsize $+$}0.46)}\\ \midrule

  \textsc{bert-qa}
&
    lo & 45.34 & \pos{({\scriptsize $+$}3.73)} & 57.89 & \pos{({\scriptsize $+$}6.15)} &  51.58 & \pos{({\scriptsize $+$}4.93)} & 58.61 & \pos{({\scriptsize $+$}11.30)} & 78.70 & \pos{({\scriptsize $+$}1.52)} & 77.87 & \pos{(\textrm{{\scriptsize $+$}2.05})} \\
+vals
    & hi & 45.21 & \pos{({\scriptsize $+$}2.41)} & 70.11 & \pos{({\scriptsize $+$}2.09)} & 59.04 & \pos{({\scriptsize $+$}2.23)} & 60.71 & \pos{({\scriptsize $+$}9.03)} & 88.45 & \nega{({\scriptsize $-$}0.15)} & 88.16 & \pos{(\textrm{{\scriptsize $\pm$}0.00})}  \\
     +drop & all & 45.28 & \pos{({\scriptsize $+$}3.11)} & 64.35 & \pos{({\scriptsize $+$}4.01)} & 55.32 & \pos{({\scriptsize $+$}3.59)}  & 59.62 & \pos{({\scriptsize $+$}10.22)} & 86.37 & \pos{({\scriptsize $+$}0.23)} & 85.91 & \pos{(\textrm{{\scriptsize $+$}0.48})}   \\\midrule

  \textsc{bert-qa} &
  lo  & 47.64 & \pos{({\scriptsize $+$}6.03)} & 57.91 & \pos{(\textbf{{\scriptsize $+$}6.17})} & 52.75 & \pos{({\scriptsize $+$}6.10)} & 58.13 & \pos{({\scriptsize $+$}10.82)} & 78.78 & \pos{(\textbf{{\scriptsize $+$}1.60})} & 77.90 & \pos{(\textbf{{\scriptsize $+$}2.08})} \\
  +vals  &
  hi  & 48.38 & \pos{({\scriptsize $+$}5.58)} & 70.48 & \pos{({\scriptsize $+$}2.46)} & 60.67 & \pos{({\scriptsize $+$}3.86)}  & 62.10 & \pos{({\scriptsize $+$}10.42)} & 88.74 & \pos{(\textbf{{\scriptsize $+$}0.14})} & 88.45 & \pos{(\textbf{{\scriptsize $+$}0.29})} \\
  +mixing &
  all & 47.99 & \pos{({\scriptsize $+$}5.82)} & 64.55 & \pos{({\scriptsize $+$}4.21)} & 56.71 & \pos{({\scriptsize $+$}4.98)}  & 60.03 & \pos{({\scriptsize $+$}10.63)} & 86.61 & \pos{(\textbf{{\scriptsize $+$}0.47})} & 86.13 & \pos{(\textbf{{\scriptsize $+$}0.70})} \\
  \midrule  

  \textsc{bert-qa} &
  lo  & 49.15 & \pos{(\textbf{{\scriptsize $+$}7.54})} & 57.89 & \pos{({\scriptsize $+$}6.15)} & 53.51 & \pos{(\textbf{{\scriptsize $+$}6.86})} & 60.18 & \pos{(\textbf{{\scriptsize $+$}12.87})} & 78.55 & \pos{({\scriptsize $+$}1.37)} & 77.74 & \pos{(\textrm{{\scriptsize $+$}1.92})} \\
  +vals &
  hi  & 50.94 & \pos{(\textbf{{\scriptsize $+$}8.14})} & 71.04 & \pos{(\textbf{{\scriptsize $+$}3.02})} & 62.10 & \pos{(\textbf{{\scriptsize $+$}5.29})} & 63.06 & \pos{(\textbf{{\scriptsize $+$}11.38})} & 88.56 & \nega{({\scriptsize $-$}0.04)} & 88.27 & \pos{(\textrm{{\scriptsize $+$}0.11})} \\
   +drop +mixing &
  all & 49.99 & \pos{(\textbf{{\scriptsize $+$}7.82})} & 64.84 & \pos{(\textbf{{\scriptsize $+$}4.50})} & 57.81 & \pos{(\textbf{{\scriptsize $+$}6.08})} & 61.56 & \pos{(\textbf{{\scriptsize $+$}12.16})} & 86.42 & \pos{({\scriptsize $+$}0.28)} & 85.96 & \pos{(\textrm{{\scriptsize $+$}0.53})} \\

\bottomrule
    \end{tabular}

    \caption{Macro and micro F$_1$ gains over \textsc{bert-qa} for 544 attributes (21,374 test examples) that took our value-based query expansion. 
%   % The attributes are \yn{divided into four groups} by median frequency and similarity to values in the training data.
    `lo' and `hi' are similarity intervals, $[0.411, 0.929)$ and $[0.929, 1.0]$, respectively.}
    \label{tab:result_stats}

\end{table*}
% except for tuples whose value is ``\textsc{null}.'' 
% As a result, we obtain 11,819 attribute-value pairs with 1,801 unique attributes and 9,317 unique values.
To convert tuples in the training set to beginning and ending positions, we tokenize both title and value, and then use matching positions if the token sequence of the value exactly matches a sub-sequence of the title. If the value matches multiple portions of the title, we use the match close to the beginning of the title. As beginning and ending positions of tuples whose value is ``\textsc{null},'' we use 0 which is a position of a \textsc{cls} token in the title. The conversion procedure is detailed in Appendix~\ref{ssec:tuples2data}.

% \subsection*{Implementation}
We implemented the above models using PyTorch~\cite{NEURIPS2019_9015} (ver.~1.7.1),
% \footnote{\url{https://github.com/pytorch/pytorch/}}
 and used ``bert-base-uncased'' in Transformers~\cite{wolf_2020} as the pre-trained \textsc{bert} (\textsc{bert}$_{\textsc{base}}$).
The implementation details and the training time are given in Appendix~\ref{ssec:detail} and Appendix~\ref{ssec:time}, respectively.

\begin{table*}[ht]
\small
    \centering
    \begin{tabular}{@{\,\,}lc@{\quad}cc@{\quad}cc@{\quad}c@{\,\,}}\toprule
% & \multicolumn{4}{c}{Seen Attributes} & \multicolumn{2}{c}{\multirow{2}{*}{Unseen Attributes}}\\
 \multirow{2}{*}{Models} & \multicolumn{2}{c}{Seen Attr. (Seen Values)} & \multicolumn{2}{c}{Seen Attr. (Unseen Values)} & \multicolumn{2}{c}{Unseen Attr.}\\ 
 % \cmidrule(lr){2-3} \cmidrule(lr){4-5} \cmidrule(lr){6-7}
 % \multicolumn{2}{c}{Unseen Values}\\
%\cmidrule(lr){2-4}
%\cmidrule(l){5-7}
& Micro F$_1$ & Macro F$_1$ & Micro F$_1$ & Macro F$_1$ & Micro F$_1$ & Macro F$_1$\\\midrule

Dictionary &
87.19 \scriptsize{($\pm$0.00)} & 83.89 \scriptsize{($\pm$0.00)} &
n/a & n/a & n/a & n/a\\

\textsc{bert-qa}~\cite{wang_2020} &
92.26 \scriptsize{($\pm$0.15)} & 73.30 \scriptsize{($\pm$2.30)} &
\underline{46.11} \scriptsize{($\pm$0.86)} & \textbf{25.43} \scriptsize{($\pm$1.24)} &
\underline{25.86} \scriptsize{($\pm$2.53)} & \underline{20.92} \scriptsize{($\pm$1.92)} \\

\textsc{bert-qa}  +vals &
\textbf{92.85} \scriptsize{($\pm$0.13)} & \textbf{86.93} \scriptsize{($\pm$0.61)} &
42.11 \scriptsize{($\pm$0.70)} & 10.98 \scriptsize{($\pm$1.01)} &
\,\,\,6.90 \scriptsize{($\pm$1.19)} & \,\,\,4.03 \scriptsize{($\pm$0.76)} \\

\textsc{bert-qa}  +vals +drop &
92.74 \scriptsize{($\pm$0.20)} & 86.21 \scriptsize{($\pm$0.58)} &
44.14 \scriptsize{($\pm$0.19)} & 16.40 \scriptsize{($\pm$0.80)} &
11.17 \scriptsize{($\pm$2.89)} & \,\,\,7.21 \scriptsize{($\pm$2.16)} \\

\textsc{bert-qa} +vals +mixing &
\underline{92.82} \scriptsize{($\pm$0.15)} & \underline{86.40} \scriptsize{($\pm$0.79)} &
45.59 \scriptsize{($\pm$0.48)} & 19.87 \scriptsize{($\pm$1.81)} &
25.39 \scriptsize{($\pm$2.63)} & 20.14 \scriptsize{($\pm$2.13)} \\

\textsc{bert-qa} +vals +drop +mixing &
92.67 \scriptsize{($\pm$0.11)} & 86.34 \scriptsize{($\pm$0.72)} &
\textbf{46.14} \scriptsize{($\pm$0.34)} & \underline{22.52} \scriptsize{($\pm$0.93)} &
\textbf{27.54} \scriptsize{($\pm$1.35)} & \textbf{21.95} \scriptsize{($\pm$1.25)} \\
\bottomrule

% \textsc{bert-qa} +vals +vdrop +mixing (3) &
% 92.93 \scriptsize{($\pm$0.15)} & 85.68 \scriptsize{($\pm$0.88)} &
% 45.96 \scriptsize{($\pm$0.90)} & 24.65 \scriptsize{($\pm$2.32)} &
% 26.29 \scriptsize{($\pm$2.41)} & 20.70 \scriptsize{($\pm$1.95)} \\
% \hline

% \textsc{bert-qa}  +3dtm +vals &
% \textbf{92.96} \scriptsize{($\pm$0.09)} & \underline{85.81} \scriptsize{($\pm$0.77)} &
% 45.85 \scriptsize{($\pm$0.67)} & 20.91 \scriptsize{($\pm$1.91)} &
% 24.50 \scriptsize{($\pm$2.26)} & 19.05 \scriptsize{($\pm$1.75)} \\

% \textsc{bert-qa}  +3dtm +vals (dropout) &
% \underline{92.93} \scriptsize{($\pm$0.15)} & 85.68 \scriptsize{($\pm$0.88)} &
% \underline{45.96} \scriptsize{($\pm$0.90)} & \underline{24.65} \scriptsize{($\pm$2.32)} &
% \textbf{26.13} \scriptsize{($\pm$2.38)} & \underline{20.60} \scriptsize{($\pm$1.90)} \\
% \hline
    \end{tabular}
    \caption{Performance on the cleaned \textsc{ae}-pub dataset in terms of the types of the attribute values; reported numbers are mean (std. dev.) of five trials. The best score is in bold face and the second best score is underlined.}
    \label{tab:performance_for_each_value_type}
\end{table*}

\begin{comment}

\begin{table}[t]
\small
\setlength{\tabcolsep}{1.3mm}
    \centering
    \begin{tabular}{l|cc|cc}\hline
    \multirow{2}{*}{Models} & \multicolumn{2}{c|}{All attributes} & \multicolumn{2}{c}{Attributes that QR ran}\\
    & Micro-F$_1$ & Macro-F$_1$ & Micro-F$_1$ & Macro-F$_1$\\\hline
    SUOpenTag &
    82.69 & 29.24 & 83.50 & 39.07\\
    \textsc{aveqa} &
    84.42 & 40.59 & 85.23 & 51.04\\
    \,\, +rand-val &
    84.52 & 42.31 & 85.32 & 53.90\\
    \,\, +freq-val &
    84.63 & \textbf{43.85} & 85.43 & \textbf{56.26}\\
    \,\, +add-vals &
    \textbf{85.05} & 42.41 & \textbf{85.86} & 54.18\\
    \,\, +avg-vals &
    \underline{84.83} & \underline{43.56} & \underline{85.63} & \underline{55.88}\\\hline
    \end{tabular}
    \caption{The performance on the cleaned \textsc{ae}-pub dataset in Table~\ref{tab:data_stats}. The second best scores are underlined. The scores under ``Attributes that QR ran'' is to see the effectiveness of the query reformulation in detail. We calculate those based on attributes whose values are seen in the training data and which takes the query reformulation. As a result, 21,550 examples for 612 attributes remain for the evaluation.}
    \label{tab:performance_on_limited_attrs}
\end{table}
\end{comment}

\subsection{Results}
\label{comparison}

% To do so, we exclude attributes that are not appeared in the training set from the evaluation because we cannot collect values for such attributes.
%
% Moreover, there are attributes that only have a ``\textsc{null}'' value in the test set.
% averaged performance if those attributes are in the test set because the recall for those attributes cannot be defined.
%
% We therefore exclude such attributes. 
% ynaga 本当は入れたい
% \footnote{Here, we exclude seen attributes whose all values are ``\textsc{null}'' in the test set, since we cannot calculate the macro recall (and F$_1$).}
%

% \subsection{Analysis}
% \label{sec:analysis}

% \footnote{When we fully use the cleaned \textsc{ae}-pub dataset, our value-driven query reformulation consistently shows better micro F$_1$ score than BERT-\textsc{qa}.}
Table~\ref{tab:performance_on_limited_attrs} shows macro\footnote{We ignored 70 attributes with only \textsc{null} since we cannot compute recall and F$_1$ for these attributes.} and micro performance of each model that are averaged over five trials.
% \footnote{\yn{We ignored attributes that have only \textsc{null} values in the test data in computing macro F$_1$, since we cannot compute recall (and F$_1$) for these attributes. The total number of such attributes is 70 (178 examples) for ``All attributes,'' and 68 (176 examples) for ``Attributes that QR ran,'' respectively.}} performance of each model.% on these 612 attributes.
%
% The total number of such attributes is 70 (178 examples) for ``All attributes,'' and 68 (176 examples) 
%
% We can see that our value-driven query expansion consistently improves the model performance.
%regardless of the ways of integrating % the attribute 
%values.
%
% We can see that the models with the query reformulation consistently show better micro F$_1$ scores than the BERT-\textsc{qa} model.
% Furthermore, the 
% The gap in the macro F$_1$ scores between the BERT-\textsc{qa} model and others is larger than the gap in the micro F$_1$ scores. These results show that our query reformulation is more effective for attributes with less examples in the test set.
% Interestingly, the model +all-vals with the best micro F$_1$ did not improve macro F$_1$. 
% The high micro precision and low micro 
The low recall of the model \textsc{bert-qa} +vals suggests that this model learns to find strings that are similar to ones retrieved from the training data (overfitting).
On the other hand, knowledge dropout and knowledge token mixing mitigates the overfitting, and improves both macro and micro F$_1$ performance.
% Overall, the model +avg-vals exhibits the stable performance in both micro and macro performance.

\paragraph{Impact
% value-based query expansion
on rare and ambiguous attributes}
% \subsection{Does query reformulation help find values for rare and inappropriate attributes?}
% Our query expansion is meant to obtain better query representations by referring to seen attribute values.
% by integrating attribute values into the query.
% This is motivated by the fact that  many attributes are rare or have 
% inappropriate 
% ambiguous names.
% the attributes 
% in the \textsc{ave} task 
% do not have enough amount of the training examples, or have inappropriate names.
% Although our \textsc{qa}-based \textsc{ave} model relies on BERT pretrained on general domain text, it is hard for the model to successfully fit the representation of rare attributes to the e-commerce domain because of less training examples.
%
% We thus examine if our method improves the performance for those attributes.
%
% According to the difference between the models BERT-\textsc{qa} and \textsc{qa} +avg-vals in F$_1$ scores, we
%We first classify attributes in Table~\ref{tab:performance_on_limited_attrs} into three groups by comparing BERT-\textsc{qa} and +avg-vals in terms of F$_1$: \textit{Improved} ($+$), \textit{Impaired} ($-$), and \textit{Tie} ($=$). 
To see if the query expansion improves the performance for rare attributes with 
% inappropriate
ambiguous names, 
we categorized the attributes that took the query expansion % method performs
% in Table~\ref{tab:performance_on_limited_attrs}
according to the number of training examples and the appropriateness of the attribute names for their values.
%
% Next, in order to see how appropriate attributes are, we compute 
% semantic similarity between attributes and their values.
% and then regard it inappropriate if the similarity is relatively low
To measure the name appropriateness, 
we exploit embeddings of the \textsc{cls} token using the \textsc{bert}$_{\textsc{base}}$ for each attribute and its seen values; when the cosine similarity between the attribute embedding and averaged value embeddings is low, we regard the attribute name as ambiguous.
% inappropriate.
% We average similarities for all values if the attribute has multiple values. 
% The similarity between attribute and value was 0.88 in average. 
We divide the attributes into four 
% into halves
according to median frequency and similarity to values. %  respectively.
% low ($[0, 0.861)$), medium ($[0.861, 0.929)$), and high ($[0.929, 1.0]$).

Table~\ref{tab:result_stats} lists % the number of +/-/= attributes 
macro and micro F$_1$ of each model and the improvements over the \textsc{bert-qa} for each category.
% 4.5の最初, Table 3のcaptionでも言及しているので消しました。
% To see the effectiveness of the query expansion, we calculate the macro F$_1$ for attributes whose values are observed in the training data and which take the query expansion.
% As a result, 21,550 examples for 612 attributes remain for the evaluation.
%
%
We can see that our 
% value-based 
query expansion tends to be more effective for attributes with low similarity.
This means
% that inappropriate attributes are likely to make vanilla \textsc{aveqa} fail extraction, and 
that the query expansion can generate more informative queries than 
% inappropriate 
ambiguous attributes alone.
Moreover, by using knowledge dropout and knowledge token mixing, we can improve macro and micro F$_1$ for rare attributes. 
These results are remarkable since the knowledge used to enhance the model comes from its training data; the model could use more parameters to solve the task itself by taking the internal knowledge induced from the training data as runtime input.

\paragraph{Impact on seen and unseen attribute values}

To see for what types of attribute values the query expansion is effective, we categorize the test examples according to the types of the training data used to solve the examples.
% and then check the performance for each category. 
% \yn{To see if the types of the knowledge referred in our query expansion affects on its performance, we categorized the test examples according the types of the referred knowledge.}
We first categorize the test examples into seen or unseen attributes. Next, we further classify the examples for the seen attributes into either seen or unseen attribute values.

Table~\ref{tab:performance_for_each_value_type} shows the performance in terms of the attribute value types. 
The query expansion improved macro F$_1$ by 13 points on the seen values for the seen attributes; these improvements were yielded by the large performance gains for rare attributes in 
Table~\ref{tab:result_stats}.
% For the seen values, the query expansion improves both micro and macro F$_1$ scores. 
% In particular, \textsc{bert-qa} +vals achieves the best micro and macro F$_1$.
% Since the improvement in the macro F$_1$ over \textsc{bert-qa} is much larger than in the micro F$_1$, the query expansion is more effective to extract values infrequently occurring in the training data. Similar results can be seen in Table~\ref{tab:result_stats}.
% This is promising because the majority of attributes have less labeled samples in the \textsc{ave} task~\cite{xu_2019,yang_2022}.
Although \textsc{bert-qa} +vals performed the best on the seen values, it performed the worst on the unseen values for the seen attributes and unseen attributes; the model is trained to match seen values in a query with a given title. Meanwhile, the two tricks enable the model to maintain the micro F$_1$ performance of \textsc{bert} on the unseen values for the seen attributes. The lower macro F$_1$ against \textsc{bert} suggests that there is still room for improvements in query representation for rare seen attributes.
% mitigate the performance drop caused by the simple use of attribute values in the query expansion.
% As expected, the knowledge token mixing is more effective for the unseen values than the knowledge dropout.
Lastly, the knowledge token mixing successfully recovered the performance of \textsc{bert} for the unseen attributes, and even improved the performance when it is used together with the knowledge dropout. 
% This means that the knowledge token mixing successfully controls the degree of dependency on values in the models.
This is possibly because the knowledge token mixing allows the model to switch its behavior for seen and unseen attributes, and the knowledge dropout strengthens the ability to induce better query representations.

% Overall, \textsc{bert-qa} with knowledge dropout and knowledge token mixing exhibits promising performance; the model shows better performance than \textsc{bert-qa} for the seen values as it is comparable to \textsc{bert-qa} for the unseen values. 

\begin{table}[t]
    \small
    \centering
    % \begin{tabular}{lp{0.5\columnwidth}p{0.85\columnwidth}cc}\toprule
    \begin{tabular}{@{}l@{\,}p{207pt}@{}}
    %\begin{tabular}{@{}l@{\,}p{500pt}@{}}
    \toprule
    
% \textbf{Context:}& storage bag fishing outdoor portable square gear tackle \textbf{container}$_{\textrm{Ours}}$ reel lure float accessories shoulder belt \textit{waterproof}$_{\textrm{\textsc{aveqa}}}$ w20\\
    % \textbf{Query:}&  use (hunting, shooting, rod, reel, mountain bikes, $\ldots$)\\
    % \midrule
    
    \textbf{C:}&
    aeronova [bicycle [\textbf{carbon mtb handlebar}]$_{\textrm{ours}}$]$_{\textsc{bert-qa}}$ mountain bikes flat handlebar mtb integrated handlebars with stem bike accessories\\
    \textbf{Q:}&
    function 1 (skiing goggles, carbon road bicycle handlebar, cycling glasses, bicycle mask, gas mask, $\ldots$)\\
    \midrule
%    \textbf{C:}&
%    robesbon [\textbf{bicycle glasses}]$_{\textrm{+vals}|\textrm{+vals (dropout)}}$ uv400 cycling glasses men women bike eyewear sports sunglasses hiking fishing running bicycle eyewear\\
%    \textbf{Q:}&
%    function 1 (skiing goggles, carbon road bicycle handlebar, cycling glasses, bicycle mask, gas mask, $\ldots$)\\
    % hot sale 15pcs 4.5x36mm night \textit{fishing float}$_{\textrm{\textsc{aveqa}}}$ chemical light fish fluorescent glowing % \textbf{fishing light stick}$_{\textrm{Ours}}$ luminous float lights\\
    % \textbf{Query:}&  type (electric guitar, monocular, dice, binoculars, helmet, $\ldots$)\\
%    \hline
    % \textbf{Cont.:}&
    % men casual \textbf{dome}$_{\mathrm{Ours}}$ \textit{baseball cap}$_{\textsc{bert-qa}}$ letters cap adjustable fashion available colors variety\\
    % \textbf{Query:}& cap type (dome)\\
    % 1 person ultra - light aluminum alloy \textit{single}$_{\textrm{\textsc{aveqa}}}$ riding tent outdoor camping tents travel portable hiking tents \textbf{double}$_{\textrm{Ours}}$ layer\\
    % \textbf{Query:}&  layers (double, single, 5 layers)\\
    % \midrule    
    \textbf{C:}&
    lfp [3.2v [\textbf{100ah}]$_{\textrm{ours}}$]$_{\textsc{bert-qa}}$ lifepo4 prismatic cell deep cycle diy lithium ion battery 72v 60v 48v 24v 100ah 200ah ev solar storage battery\\
    \textbf{Q:}& nominal capacity (14ah, 40ah, 17.4ah)\\
    % 1 person ultra - light aluminum alloy \textit{single}$_{\textrm{\textsc{aveqa}}}$ riding tent outdoor camping tents travel portable hiking tents \textbf{double}$_{\textrm{Ours}}$ layer\\
    % \textbf{Query:}&  layers (double, single, 5 layers)\\
    \midrule    
    \textbf{C:}&
    camel outdoor softshell [\textbf{men}]$_{\textsc{bert-qa}}$'s hiking jacket windproof thermal jacket for [camping]$_{\textrm{ours}}$ ski thick warm coats\\
    \textbf{Q:} & suitable (men, camping, kids, saltwater/freshwater, women, 4-15y, mtb cycling shoes, $\ldots$)\\
    % aldomour 2018 hot men fashion warm winter men shoes autumn \textit{leather}$_{\textrm{Ours}}$ footwear for new high top \textbf{canvas}$_{\textrm{\textsc{aveqa}}}$ casual shoes men wjp\\
    % \textbf{Query:} & upper material (canvas, leather, pu, genuine leather, satin, $\ldots$)\\
    \bottomrule
    \end{tabular}
    \caption{Example outputs of \textsc{bert-qa} with and without query expansion for given C(ontext) and Q(uery).}
    \label{tab:examples}
\end{table}

\paragraph{Example outputs} Table~\ref{tab:examples} shows examples of the actual model outputs for a given context and query (attribute (seen values)). 
%actual model prediction.
%
In the first two examples, \textit{function 1} and \textit{nominal capacity} are ambiguous and rare attributes, respectively, and are thereby hard for the \textsc{bert-qa} to extract \textbf{correct} values without the help of our query expansion. 
% The last example shows the wrong prediction of the model with the query expansion. 
As shown in the last example, when there are more than one candidates as values of a given attribute, our query expansion is still unstable.

\section{Conclusions}

We have proposed simple query expansion based on possible values of a given query (attribute) for \textsc{qa}-based attribute extraction. % first
With the two tricks to mimic the imperfection of the value knowledge, we
% first
retrieve values of given attributes from the training data, and then use the obtained values as knowledge to induce better query representations.
Experimental results on our cleaned version of the public AliExpress dataset demonstrate that our method improves the performance of product attribute extraction, especially for rare and ambiguous attributes.
% with 
% ambiguous 
% inappropriate 
% names.
%

% Although we focus on seen attributes, we
We will leverage external resources to handle unseen attributes (preliminary experiments are shown in Appendix~\ref{ssec:external}). We will release the script to build our cleaned \textsc{ae}-pub dataset.\footnote{\url{http://www.tkl.iis.u-tokyo.ac.jp/~ynaga/acl2022/}} 

% We will release the cleaned \textsc{ae}-pub datasets 
% (\S~\ref{ssec:dataset}) 
% to promote the reproducibility of our results. 
% ynaga
% As future work, motivated by \textsc{realm}~\cite{pmlr-v119-guu20a}, we plan to leverage contexts of attribute values to realize a better value-driven query expansion.

% more sophisticated approaches to capture variations in attribute values into the query reformulation.


\section*{Acknowledgements}

This work (second author) was partially supported by JSPS KAKENHI Grant Number 21H03494.
We thank the anonymous reviewers for their hard work.

\bibliographystyle{acl_natbib}
\bibliography{skeiji}

% new page 入れろと書いていないので
%\newpage

\appendix
% \section*{Appendix}

\section{Appendix}

\begin{table*}[t]
\small
    \centering
    \begin{tabular}{@{\,\,}lc@{\;\;\;}c@{\;\;\;}cc@{\;\;\;}c@{\;\;\;}c@{\,\,}}\toprule
\multirow{2}{*}{Models} & \multicolumn{3}{c}{Macro} &  \multicolumn{3}{c}{Micro}\\
& P (\%) & R (\%) & F$_1$ & P (\%) & R (\%) & F$_1$\\
\midrule

\multicolumn{7}{@{\,\,}c@{\,\,}}{Seen attributes (seen values)}\\
\midrule
\textsc{bert-qa}~\cite{wang_2020}
 & 73.10 \scriptsize{($\pm$3.99)} & 66.86 \scriptsize{($\pm$2.97)} & 69.83 \scriptsize{($\pm$3.42)}
 & 95.50 \scriptsize{($\pm$0.13)} & 92.14 \scriptsize{($\pm$0.21)} & 93.79 \scriptsize{($\pm$0.10)} \\
%\midrule
\multicolumn{7}{@{\,\,}l@{\,\,}}{\textit{w/ values in the training  data}}\\
\,\,\,\textsc{bert-qa} +vals
 & \textbf{87.66} \scriptsize{($\pm$1.21)} & \textbf{82.60} \scriptsize{($\pm$1.13)} & \textbf{85.06} \scriptsize{($\pm$1.15)}
 & 95.76 \scriptsize{($\pm$0.21)} & 92.54 \scriptsize{($\pm$0.12)} & 94.13 \scriptsize{($\pm$0.15)} \\
\,\,\,\textsc{bert-qa} +vals +drop
 & 87.28 \scriptsize{($\pm$0.65)} & 81.83 \scriptsize{($\pm$0.90)} & 84.47 \scriptsize{($\pm$0.70)}
 & \textbf{95.84} \scriptsize{($\pm$0.19)} & 92.81 \scriptsize{($\pm$0.36)} & 94.30 \scriptsize{($\pm$0.17)} \\
\,\,\,\textsc{bert-qa} +vals +mixing 
 & 86.98 \scriptsize{($\pm$0.91)} & 81.36 \scriptsize{($\pm$1.13)} & 84.07 \scriptsize{($\pm$0.88)}
 & \textbf{95.84} \scriptsize{($\pm$0.16)} & 92.80 \scriptsize{($\pm$0.25)} & 94.29 \scriptsize{($\pm$0.10)} \\
\,\,\,\textsc{bert-qa} +vals +drop +mixing 
 & 86.43 \scriptsize{($\pm$0.98)} & 81.48 \scriptsize{($\pm$0.70)} & 83.88 \scriptsize{($\pm$0.78)}
 & 95.79 \scriptsize{($\pm$0.20)} & \textbf{93.12} \scriptsize{($\pm$0.11)} & \textbf{94.44} \scriptsize{($\pm$0.15)} \\
 
%\midrule
\multicolumn{7}{@{\,\,}l@{\,\,}}{\textit{w/ values in the training and development data}}\\
%\textsc{bert-qa}
% & 73.10 \scriptsize{($\pm$3.99)} & 66.86 \scriptsize{($\pm$2.97)} & 69.83 \scriptsize{($\pm$3.42)}
% & 95.50 \scriptsize{($\pm$0.13)} & 92.14 \scriptsize{($\pm$0.21)} & 93.79 \scriptsize{($\pm$0.10)} \\
\,\,\,\textsc{bert-qa} +vals
 & \nega{86.71} \scriptsize{($\pm$1.14)} & \nega{81.50} \scriptsize{($\pm$0.83)} & \nega{84.02} \scriptsize{($\pm$0.96)}
 & \nega{95.44} \scriptsize{($\pm$0.23)} & \nega{92.00} \scriptsize{($\pm$0.13)} & \nega{93.69} \scriptsize{($\pm$0.18)} \\
\,\,\,\textsc{bert-qa} +vals +drop
 & \nega{85.29} \scriptsize{($\pm$1.04)} & \nega{80.29} \scriptsize{($\pm$0.94)} & \nega{82.71} \scriptsize{($\pm$0.93)}
 & \nega{95.44} \scriptsize{($\pm$0.20)} & \nega{92.44} \scriptsize{($\pm$0.31)} & \nega{93.92} \scriptsize{($\pm$0.12)} \\
\,\,\,\textsc{bert-qa} +vals +mixing 
 & \nega{85.89} \scriptsize{($\pm$1.50)} & \nega{80.51} \scriptsize{($\pm$2.26)} & \nega{83.11} \scriptsize{($\pm$1.85)}
 & \nega{95.77} \scriptsize{($\pm$0.16)} & \nega{92.77} \scriptsize{($\pm$0.22)} & \nega{94.25} \scriptsize{($\pm$0.08)} \\
\,\,\, \textsc{bert-qa} +vals +drop +mixing 
& \nega{85.65} \scriptsize{($\pm$0.57)} & \nega{80.72} \scriptsize{($\pm$0.69)} & \nega{83.11} \scriptsize{($\pm$0.56)}
 & \nega{95.75} \scriptsize{($\pm$0.16)} & \nega{93.06} \scriptsize{($\pm$0.12)} & \nega{94.39} \scriptsize{($\pm$0.12)} \\
 
\midrule
\multicolumn{7}{@{\,\,}c@{\,\,}}{Seen attributes (unseen values)}\\
\midrule

%\textsc{bert-qa}  +vals　& 54.67 \scriptsize{($\pm$0.55)} & 47.90 \scriptsize{($\pm$0.68)} & 51.06 \scriptsize{($\pm$0.53)}　& 87.61 \scriptsize{($\pm$0.21)} & 81.69 \scriptsize{($\pm$0.20)} & 84.55 \scriptsize{($\pm$0.16)}\\

% \textsc{bert-qa}  +vals +drop & 56.81 \scriptsize{($\pm$0.95)} & 51.00 \scriptsize{($\pm$0.61)} & 53.75 \scriptsize{($\pm$0.74)} & 87.43 \scriptsize{($\pm$0.31)} & 81.86 \scriptsize{($\pm$0.43)} & 84.55 \scriptsize{($\pm$0.15)}\\

% \textsc{bert-qa} +vals +mixing & 57.27 \scriptsize{($\pm$0.69)} & 51.30 \scriptsize{($\pm$0.75)} & 54.12 \scriptsize{($\pm$0.60)} & \textbf{87.76} \scriptsize{($\pm$0.78)} & 82.12 \scriptsize{($\pm$0.34)} & \textbf{84.84} \scriptsize{($\pm$0.31)}\\

% \textsc{bert-qa} +vals +drop +mixing & \textbf{58.26} \scriptsize{($\pm$0.33)} & \textbf{52.93} \scriptsize{($\pm$0.86)} & \textbf{55.46} \scriptsize{($\pm$0.57)} & 87.20 \scriptsize{($\pm$0.37)} & \textbf{82.18} \scriptsize{($\pm$0.11)} & 84.62 \scriptsize{($\pm$0.20)}\\
\textsc{bert-qa}~\cite{wang_2020}
 & \textbf{29.72} \scriptsize{($\pm$2.20)} & \textbf{24.72} \scriptsize{($\pm$1.58)} & \textbf{26.99} \scriptsize{($\pm$1.83)}
 & 34.44 \scriptsize{($\pm$3.47)} & \textbf{21.28} \scriptsize{($\pm$1.80)} & \textbf{26.28} \scriptsize{($\pm$2.26)} \\
% \midrule
 \multicolumn{7}{@{\,\,}l@{\,\,}}{\textit{w/ values in the training  data}}\\
\,\,\,\textsc{bert-qa} +vals
 & 16.89 \scriptsize{($\pm$1.49)} & 12.94 \scriptsize{($\pm$1.46)} & 14.65 \scriptsize{($\pm$1.48)}
 & 31.56 \scriptsize{($\pm$2.40)} & 12.42 \scriptsize{($\pm$1.15)} & 17.83 \scriptsize{($\pm$1.55)} \\
\,\,\,\textsc{bert-qa} +vals +drop
 & 22.32 \scriptsize{($\pm$1.48)} & 18.77 \scriptsize{($\pm$1.06)} & 20.39 \scriptsize{($\pm$1.24)}
 & 37.06 \scriptsize{($\pm$1.09)} & 16.99 \scriptsize{($\pm$0.70)} & 23.30 \scriptsize{($\pm$0.81)} \\
\,\,\,\textsc{bert-qa} +vals +mixing 
 & 24.10 \scriptsize{($\pm$1.45)} & 18.98 \scriptsize{($\pm$0.90)} & 21.23 \scriptsize{($\pm$1.09)}
 & 35.07 \scriptsize{($\pm$1.51)} & 16.77 \scriptsize{($\pm$1.02)} & 22.68 \scriptsize{($\pm$1.22)} \\
\,\,\,\textsc{bert-qa} +vals +drop +mixing 
 & 27.19 \scriptsize{($\pm$1.20)} & 22.31 \scriptsize{($\pm$1.00)} & 24.51 \scriptsize{($\pm$1.06)}
 & 36.60 \scriptsize{($\pm$0.50)} & 18.33 \scriptsize{($\pm$0.73)} & 24.42 \scriptsize{($\pm$0.64)} \\
 
%\midrule
\multicolumn{7}{@{\,\,}l@{\,\,}}{\textit{w/ values in the training and development data}}\\
% \textsc{bert-qa} & \textbf{29.72} \scriptsize{($\pm$2.20)} & \textbf{24.72} \scriptsize{($\pm$1.58)} & \textbf{26.99} \scriptsize{($\pm$1.83)} & 34.44 \scriptsize{($\pm$3.47)} & \textbf{21.28} \scriptsize{($\pm$1.80)} & \textbf{26.28} \scriptsize{($\pm$2.26)} \\
\,\,\,\textsc{bert-qa} +vals
 & \pos{24.03} \scriptsize{($\pm$1.81)} & \pos{17.94 }\scriptsize{($\pm$1.56)} & \pos{20.54} \scriptsize{($\pm$1.68)}
 & \pos{36.54} \scriptsize{($\pm$2.25)} & \pos{15.71} \scriptsize{($\pm$1.31)} & \pos{21.97} \scriptsize{($\pm$1.67)} \\
\,\,\,\textsc{bert-qa} +vals +drop
 & \pos{27.27} \scriptsize{($\pm$1.09)} & \pos{22.30} \scriptsize{($\pm$1.24)} & \pos{24.53} \scriptsize{($\pm$1.17)}
 & \pos{\textbf{39.49}} \scriptsize{($\pm$2.31)} & \pos{19.16} \scriptsize{($\pm$0.59)} & \pos{25.79} \scriptsize{($\pm$0.91)} \\
\,\,\,\textsc{bert-qa} +vals +mixing
 & \pos{27.52} \scriptsize{($\pm$1.02)} & \pos{21.56} \scriptsize{($\pm$1.02)} & \pos{24.17} \scriptsize{($\pm$0.97)}
 & \pos{37.35} \scriptsize{($\pm$1.02)} & \pos{18.77} \scriptsize{($\pm$0.97)} & \pos{24.98} \scriptsize{($\pm$1.02)} \\
\,\,\,\textsc{bert-qa} +vals +drop +mixing
 & \pos{28.57} \scriptsize{($\pm$1.11)} & \pos{23.44} \scriptsize{($\pm$1.13)} & \pos{25.75} \scriptsize{($\pm$1.12)}
 & \pos{37.64} \scriptsize{($\pm$0.96)} & \pos{19.67} \scriptsize{($\pm$0.60)} & \pos{25.83} \scriptsize{($\pm$0.67)} \\
 
% \textsc{bert-qa} & 54.25 \scriptsize{($\pm$0.76)} & 48.14 \scriptsize{($\pm$0.75)} & 51.01 \scriptsize{($\pm$0.74)} & 86.99 \scriptsize{($\pm$0.30)} & 81.08 \scriptsize{($\pm$0.16)} & 83.93 \scriptsize{($\pm$0.18)}\\

% \textsc{bert-qa}  +vals +drop & 56.05 \scriptsize{($\pm$0.67)} & 50.52 \scriptsize{($\pm$0.57)} & 53.14 \scriptsize{($\pm$0.61)} & 86.83 \scriptsize{($\pm$0.43)} & 81.48 \scriptsize{($\pm$0.42)} & 84.07 \scriptsize{($\pm$0.18)} \\

% \textsc{bert-qa} +vals +mixing & 56.90 \scriptsize{($\pm$0.50)} & 51.26 \scriptsize{($\pm$0.97)} & 53.93 \scriptsize{($\pm$0.56)} & 87.50 \scriptsize{($\pm$0.77)} & 82.07 \scriptsize{($\pm$0.36)} & 84.70 \scriptsize{($\pm$0.32)}\\

% \textsc{bert-qa} +vals +drop +mixing & 57.26 \scriptsize{($\pm$0.56)} & 52.36 \scriptsize{($\pm$0.49)} & 54.70 \scriptsize{($\pm$0.34)} & 86.87 \scriptsize{($\pm$0.37)} & 82.10 \scriptsize{($\pm$0.12)} & 84.42 \scriptsize{($\pm$0.19)}\\
\midrule

\multicolumn{7}{@{\,\,}c@{\,\,}}{Unseen attributes}\\
\midrule
\textsc{bert-qa}~\cite{wang_2020}
 & 42.22 \scriptsize{($\pm$6.67)} & 42.22 \scriptsize{($\pm$6.67)} & 42.22 \scriptsize{($\pm$6.67)}
 & 59.23 \scriptsize{($\pm$8.78)} & 45.26 \scriptsize{($\pm$6.32)} & 51.22 \scriptsize{($\pm$7.14)} \\
% \midrule
\multicolumn{7}{@{\,\,}l@{\,\,}}{\textit{w/ values in the training data}}\\
\,\,\,\textsc{bert-qa} +vals
 & 15.56 \scriptsize{($\pm$2.22)} & 15.56 \scriptsize{($\pm$2.22)} & 15.56 \scriptsize{($\pm$2.22)}
 & 64.00 \scriptsize{($\pm$9.70)} & 14.74 \scriptsize{($\pm$2.11)} & 23.91 \scriptsize{($\pm$3.30)} \\
\,\,\,\textsc{bert-qa} +vals +drop
 & 19.44 \scriptsize{($\pm$2.48)} & 18.33 \scriptsize{($\pm$1.36)} & 18.85 \scriptsize{($\pm$1.85)}
 & 56.67 \scriptsize{($\pm$9.33)} & 18.95 \scriptsize{($\pm$2.58)} & 28.37 \scriptsize{($\pm$3.98)} \\
\,\,\,\textsc{bert-qa} +vals +mixing
 & 42.22 \scriptsize{($\pm$4.44)} & 42.22 \scriptsize{($\pm$4.44)} & 42.22 \scriptsize{($\pm$4.44)}
 & 61.69 \scriptsize{($\pm$3.15)} & 45.26 \scriptsize{($\pm$4.21)} & 52.03 \scriptsize{($\pm$2.92)} \\
\,\,\,\textsc{bert-qa} +vals +drop +mixing
 & 42.22 \scriptsize{($\pm$2.72)} & 42.22 \scriptsize{($\pm$2.72)} & 42.22 \scriptsize{($\pm$2.72)}
 & 54.42 \scriptsize{($\pm$2.48)} & 45.26 \scriptsize{($\pm$2.58)} & 49.41 \scriptsize{($\pm$2.51)} \\

%\midrule
\multicolumn{7}{@{\,\,}l@{\,\,}}{\textit{w/ values in the training and development data}}\\
%\textsc{bert-qa}
% & 42.22 \scriptsize{($\pm$6.67)} & 42.22 \scriptsize{($\pm$6.67)} & 42.22 \scriptsize{($\pm$6.67)}
% & 59.23 \scriptsize{($\pm$8.78)} & 45.26 \scriptsize{($\pm$6.32)} & 51.22 \scriptsize{($\pm$7.14)} \\
\,\,\,\textsc{bert-qa}  +vals
 & \pos{37.78} \scriptsize{($\pm$2.22)} & \pos{37.78} \scriptsize{($\pm$2.22)} & \pos{37.78} \scriptsize{($\pm$2.22)}
 & \pos{69.33} \scriptsize{($\pm$1.33)} & \pos{35.79} \scriptsize{($\pm$2.11)} & \pos{47.19} \scriptsize{($\pm$2.17)} \\
\,\,\,\textsc{bert-qa}  +vals +drop
 & \pos{43.33} \scriptsize{($\pm$2.22)} & \pos{43.33} \scriptsize{($\pm$2.22)} & \pos{43.33} \scriptsize{($\pm$2.22)}
 & \pos{72.18} \scriptsize{($\pm$1.09)} & \pos{41.05} \scriptsize{($\pm$2.11)} & \pos{52.32} \scriptsize{($\pm$2.02)} \\
\,\,\,\textsc{bert-qa} +vals +mixing
 & \pos{50.00} \scriptsize{($\pm$3.51)} & \pos{50.00} \scriptsize{($\pm$3.51)} & \pos{50.00} \scriptsize{($\pm$3.51)}
 & \pos{\textbf{74.67}} \scriptsize{($\pm$2.88)} & \pos{52.63} \scriptsize{($\pm$3.33)} & \pos{61.69} \scriptsize{($\pm$2.86)} \\
\,\,\,\textsc{bert-qa} +vals +drop +mixing
 & \pos{\textbf{52.22}} \scriptsize{($\pm$2.72)} & \pos{\textbf{52.22}} \scriptsize{($\pm$2.72)} & \pos{\textbf{52.22}} \scriptsize{($\pm$2.72)}
 & \pos{73.38} \pos{\scriptsize{($\pm$4.20)}} & \pos{\textbf{54.74}} \scriptsize{($\pm$2.58)} & \pos{\textbf{62.66}} \scriptsize{($\pm$2.80)} \\

\bottomrule
    \end{tabular}
    \caption{Performance on seen and unseen attributes in Table~\ref{tab:performance_for_each_value_type} whose new values are retrieved from the development data and are used for the query expansion; reported numbers are mean (std.\ dev.) of five trials.}
    \label{tab:external}
\end{table*}

\subsection{How to Convert Tuples to Labeled Data}
\label{ssec:tuples2data}
Let’s say, we have a tuple of $\langle$product title, attribute, value$\rangle = \langle$\textit{golf clubs putter pu neutral golf grip, material, pu}$\rangle$, and try to obtain beginning and ending positions of the value in the title. First, we tokenize both title and value using BertTokenizer, and then find a partial token sequence of the title that exactly matches with the token sequence of the value. By performing the match over the tokenization results, we can avoid matching a part of tokens in the title to the value. In case of this example, we can prevent the value \textit{pu} from matching to the first two characters of \textit{putter}. As a result, the value \textit{pu} matches to the token \textit{pu} in the title, and we properly obtain the beginning and ending positions of \textit{pu} in the title.

\subsection{Implementation Details}
\label{ssec:detail}
We implemented all the models used in our experiments using PyTorch~\cite{NEURIPS2019_9015} (ver.~1.7.1),\footnote{\url{https://github.com/pytorch/pytorch/}}
 and used ``bert-base-uncased'' in Transformers~\cite{wolf_2020}\footnote{\label{note2}\url{https://huggingface.co/models}}  as the pre-trained \textsc{bert} (\textsc{bert}$_{\textsc{base}}$).
The dimension of the hidden states ($D$) is 768, and
the maximum token length of the product title is 64.
We set the maximum token length of the query to 32 for \kj{all models with the exception of} models with the query expansion. To make as many attribute values as possible,
% long as the memory is available, 
we set 192 to the maximum token length of the query for the models using the query expansion, and truncate the concatenated string if the length exceeds 192. 
We set a rate of dropout over values to 0.2. The total number of parameters in \textsc{bert-qa} with our query expansion is 109M. We train the models five times with varying random seeds, and average the results.

%

% When training models with knowledge token mixing, for each training example, we construct two queries from the seen attribute, one with \textsc{seen} token prepended and its seen values postpended, and the other with \textsc{unseen} token prepended, and then put them with the title into the same mini-batch.

Regarding to \textsc{aveqa}, the loss of the distilled masked language model got NaN if we followed the algorithm in the paper. We instead used BERTMLMHead class implemented in Transformers.\footnoteref{note2}

We use Adam~\cite{adam} with a learning rate of $10^{-5}$ as the optimizer.
% As a set of values for each attribute, 
We trained the models up to 20 epochs with a batch size of 32 and chose the models that perform the best micro F$_1$ on the development set for the \kj{test set evaluation}.

\subsection{Training Time}
\label{ssec:time}
We used an NVIDIA Quadro M6000 GPU on a server with an
Intel{\small\textregistered}~Xeon{\small\textregistered} E5-2643 v4
3.40GHz CPU with 512GB main memory for training. It took around two hours per epoch for training \textsc{bert-qa} with our query expansion, while it took around 25 minutes per epoch for training the \textsc{bert-qa}.

\subsection{Preliminary experiments using external resource to obtain the value knowledge}
\label{ssec:external}
As we have discussed in \S~\ref{ssec:idea}, we can utilize external resource other than the training data of the model to perform the query expansion. We here evaluate the \textsc{bert}-\textsc{qa} models that have been already trained with our query expansion, using the development data as external (additional) resource to obtain the value knowledge in testing. If new values are retrieved from the development data, the models will build longer queries for attributes. We here evaluate such attributes with longer queries among the seen and unseen attributes in Table~\ref{tab:performance_for_each_value_type}.

Table~\ref{tab:external} shows the performance of the \textsc{bert}-\textsc{qa} models with our query expansion on 288 seen values for 107 seen attributes, 339 unseen values for 131 seen attributes, and 19 values for 18 unseen attributes, for which new values are retrieved from the development data.
We can observe that the new values retrieved from the development data boosted the performance of the \textsc{bert}-\textsc{qa} models with our query expansion on the unseen values for the seen attributes and the unseen attributes, whereas they did not increase the performance on the seen values for the seen attributes. In the future, we will explore a better way to leverage the value knowledge in the external resources other than the training data of the \textsc{qa}-based models.

\end{document}